\documentclass[conference,compsoc]{IEEEtran}
\ifCLASSOPTIONcompsoc
  \usepackage[nocompress]{cite}
\else
  \usepackage{cite}
\fi
\ifCLASSINFOpdf
  \usepackage[pdftex]{graphicx}
\else
\fi
\usepackage[cmex10]{amsmath}
\usepackage{algorithm,algpseudocode}

\usepackage{subfigure}

\begin{document}
%
\title{Semantic segmentation of trajectories with agent models}

\author{
\IEEEauthorblockN{Daisuke Ogawa, Toru Tamaki, Bisser Raytchev, Kazufumi Kaneda}
\IEEEauthorblockA{
Hiroshima University\\
Higashi-Hiroshima, Hiroshima 739-8527 Japan
}
}


%


\maketitle

\begin{abstract}

In many cases, such as trajectories clustering and classification, we often divide a trajectory into segments as  preprocessing.
In this paper, we propose a trajectory semantic segmentation method based on learned behavior models.
In the proposed method, we learn some behavior models from video sequences.
Next, using learned behavior models and a hidden Markov model, we segment a trajectory into semantic segments.
Comparing with the Ramer-Douglas-Peucker algorithm, we show the effectiveness of the proposed method.

\end{abstract}


%
\IEEEpeerreviewmaketitle

\section{Introduction}

Analyzing behavior and trajectories of pedestrians captured by video cameras are
one of important topics in computer vision, which has been widely studied over the decades.
In such studies, segmentation of trajectories often is performed for reducing computation cost
and extracting local information. There are three typical approaches \cite{Zheng}:
\begin{itemize}
 \item Temporal segmentation: splitting a trajectory at points where two observed locations are temporally away from each other .
 \item Shape-based segmentation: splitting at points of larger curvature indicating that the target may change its direction at that point. This is used for simplifying the shape of trajectories, and the Ramer-Douglas-Peucker algorithm \cite{Ramer,Douglas-Peucker} is a famous approach.
 \item Semantic segmentation: dividing a whole trajectory into semantically meaningful segments, and many methods have been proposed for different tasks \cite{Yuan,Lee,Zheng2008a,Zheng2008b,Zheng2010}.
\end{itemize}
In this paper we focus on the third type, semantic segmentation, of trajectories based on models of human behavior (or agents). 
It would be very beneficial if we could have segments related to behavior, however no segmentation methods of trajectories have been proposed for the task of human behavior analysis.
Our proposed method first estimates agent models by Mixture model of Dynamic pedestrian Agents (MDA) \cite{MDA}, then segment trajectories with the learned agent models by using Hidden Markov Models (HMM) \cite{Baum,Viterbi}.

\section{Related work}

The Ramer-Douglas-Peucker algorithm \cite{Ramer,Douglas-Peucker} is often used for
trajectory simplification. It segments a trajectory while preserving important points in order to
keep the trajectory shape as much as possible. First the start and end points of a trajectory are kept.
Next it finds the most far point away from the line between two kept points,
and keep it if the distance is larger than threshold $\epsilon$.
This process iterates recursively until no further points are kept.
Finally, all of the kept points are used to segment the trajectory.
This method is simple and preserve the rough shape of the trajectory,
while an appropriate value of $\epsilon$ has to be specified.

Task oriented methods are also proposed.
Yuan et al. \cite{Yuan} propose a system called T-Finder, which recommends
to taxi drivers places where as many potential customers exist as possible,
and to end users places where taxis are expected to be find.
To this end, they estimate taxi locations based on taxi driving trajectories
and segments the trajectories as pre-processing.
Lee et al. \cite{Lee} proposes TRAOD, an algorithm for finding outliers in trajectories
based on segmentation by using the Minimum Description Length (MDL) principle.
Zheng et al. estimates Transportation Mode \cite{Zheng2008a,Zheng2008b,Zheng2010}
such as walk, car, bus, and bike used for semantic segmentation in terms of a mode of transportation.

In contrast, our proposed method uses semantic human behavior models, called agent models,
learned from pedestrian trajectories in videos.

\section{Mixture model of Dynamic pedestrian Agents}
\label{subsec:MDA}

In this section, we briefly describe 
Mixture model of Dynamic pedestrian Agents (MDA) \cite{MDA} proposed by Zhou et al.
for learning behavior models, or agents.
MDA is a hierarchical Bayesian model that represents
pedestrian trajectories by a mixture model of dynamics and belief.
By a modified Kalman filter handling missing observations in trajectories,
and an iterative EM algorithm, parameters of dynamics and belief of each agents
are estimated.
Finally, trajectories are clustered based on the estimated agents.
In our proposed method, we use the estimated agents for segmentation of trajectories, instead of clustering.

\def\y{\boldsymbol{y}}
\def\x{\boldsymbol{x}}
\def\b{\boldsymbol{b}}
\let\oldmu\mu
\def\mu{\boldsymbol{\oldmu}}

\subsection{Formulation}

Let $\y_t \in R^2$ be two-dimensional coordinates of a pedestrian at time $t$,
and $\x_t \in R^2$ be the corresponding state
of the following linear dynamic system;
\begin{align}
\x_t &\sim P(\x_t | \x_{t-1}) = N(\x_t | A \x_{t-1} + \b, Q) \label{eq:xt}
\\
\y_t &\sim P(\y_t | \x_t) = N(\y_t | \x_t, R) \label{eq:yt},
\end{align}
where 
$N(\cdot)$ represents a normal distribution with covariance matrices $Q, R \in R^{2\times 2}$,
and $A \in R^{2\times 2}$ is state transition matrix and $\b \in R^2$ is translation vector,
assuming the state transition be a similar transformation.
In this paper we use explicitly use the translation vector for similar transformation,
while Zhou et la. \cite{MDA} used homogeneous coordinates for their formulation.

MDA represents pedestrian trajectories by a mixture of 
dynamics $D$ and belief $B$.
Here, dynamics $D = (A, \b, Q, R)$ describes dynamics of human movement in the two-dimensional scene.
Belief $B$ describes the starting point $\x_s$ and end point $\x_e$ of the trajectory,
each represented by normal distributions as follows;
\begin{align}
\x_s &\sim p(\x_s) = N(\x_s | \mu_s, \Phi_s)
\\
\x_e &\sim p(\x_e) = N(\x_e | \mu_e, \Phi_e).
\end{align}
That is, belief is represented as $B = (\mu_s, \Phi_s, \mu_e, \Phi_e)$,
describing where it starts and to where it is going.
The mixture weights are written as $\pi_m = p(z = m)$
where hidden variable $z$ represents that the trajectory is generated by agent $m$.

\begin{figure}[tb]
\begin{center}
\includegraphics[width=0.8\linewidth]{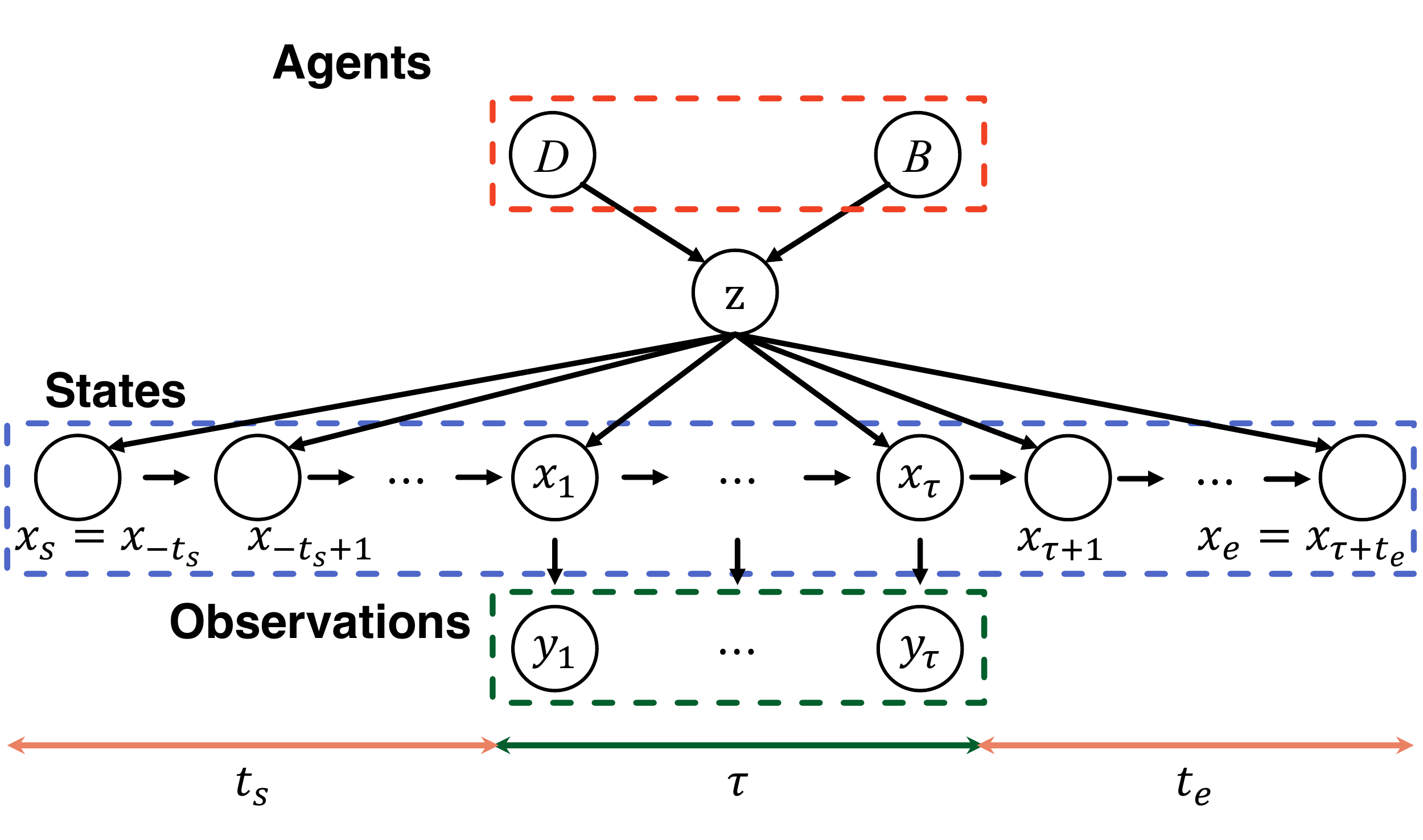}
\end{center}
\caption{State space model of MDA \cite{MDA}}
\label{fig:1}
\end{figure}

Furthermore, observation $\y = \{ \y_1, \y_2, \ldots, \y_\tau\}$ of length $\tau$
is assumed not to start and end at the exact start and end points 
$\x_s$ and $\x_e$ of the agent, that is, stats of the trajectory exists
before and after the observed points of the trajectory;
\begin{align}
\x = \{ 
& \x_s = \x_{-t_s}, \x_{-t_s + 1}, \ldots, \x_0, \notag\\
& \x_1, \x_2, \ldots, \x_\tau, \notag\\
& \x_{\tau + 1}, \ldots, \x_{\tau + t_e} = \x_e
\}.
\end{align}
Hereafter, $\x_{1:T}$ denotes the sequence of states $\x$ except $\x_s$ and $\x_e$.
Figure \ref{fig:1} shows the state space model of MDA.

\subsection{Learning}

Given $K$ trajectories $Y = \{ \y^k \}$
MDA estimates $M$ agents $\Theta = \{ (D_m, B_m, \pi_m)\}$
by maximizing the following log likelihood;
\begin{align}
L = \sum_k \log p(\y^k | \x^k, z^k, t_s^k, t_e^k,\Theta)
\end{align}
This can be rewritten by replacing 
hidden variables $Z = \{ z^k \}, T = \{ (t_s^k, t_e^k) \}$
with $H = \{Z, T\}, h^k = \{ z^k, t_s^k, t_e^k \}$
as follows;
\begin{align}
L = \sum_k \log p(\y^k | \x^k, h^k, \Theta)
\end{align}
The EM algorithm estimates iteratively as $H$ is not observed.

\subsubsection{E step}

\begin{align}
Q(\Theta, \hat\Theta)
&= E_{X, H | Y, \hat\Theta} [L]
\\
&= E_{H | \hat{X}, Y, \hat\Theta} [ E_{X | Y, H, \hat\Theta} [L] ]
\\
&= \sum_k \sum_{h^k}
\gamma^k E_{\x^k | \y^k, h^k} [L]
\end{align}
Hereafter, 
$E_{\x^k | \y^k, h^k} [\x^k] = \hat\x^k$ is denoted as $\x^k$,
which is computed by the modified Kalman filter \cite{Palma2007}.

Weights are given as follows;
\begin{align}
\gamma^k
&=
\frac{p( h^k | \hat\x_{1:T}^k, \hat\Theta) 
      p( \y^k| h^k, \hat\x_{1:T}^k, \hat\Theta)
      p( \hat\x_s^k, \hat\x_e^k | \hat\Theta)}
      {p( \y^k | \hat\x_{1:T}^k, \hat\Theta)
       p( \hat\x_s^k, \hat\x_e^k | \hat\Theta)}
\end{align}
Note that we assume the conditional independence between $\y^k$ and $\hat\x_s^k, \hat\x_e^k$,
and between $\hat\x_s^k, \hat\x_e^k$ and $ h^k, \hat\x_{1:T}^k$.

By further assuming the independence among hidden variables $z, t_s, t_e$
and the conditional independence between$\hat\x^k$ and $\hat\Theta$,
we have 
\begin{align}
p(h^k | \hat\x^k, \hat\Theta) = p(h^k) = p(z^k, t_s^k, t_e^k) = p(z^k) p(t_s^k) p(t_e^k)
\end{align}
By removing $t_s, t_e$ by assuming those be uniform,
we have
\begin{align}
\gamma^k
&= \frac{p(z^k) p( \y^k | h^k, \hat\x_{1:T}^k, \hat\Theta)
         p(\hat\x_s^k)
         p(\hat\x_e^k)}
{\sum_{h^k} p(z^k) p( \y^k | h^k, \hat\x_{1:T}^k, \hat\Theta)
                   p(\hat\x_s^k)
                   p(\hat\x_e^k)}
\end{align}
where $p( \y^k | h^k, \hat\x_{1:T}^k, \hat\Theta)$ is computed 
by the modified Kalman filter \cite{Palma2007} considering
unobserved states 
$\{ \x_{-t_s}, \x_{-t_s + 1}, \ldots, \x_0, \x_{\tau + 1}, \ldots, \x_{\tau + t_e} \}$.

\subsubsection{M step}

Next we find 
$\hat\Theta = \mathrm{arg}\max_\Theta Q(\Theta, \hat\Theta)$
by solving a system of equations
obtained by differentiating $Q$ with respect to $\Theta$,
resulting in the following analytical solutions;

\subsubsection{Implementation}

In summary, the EM algorithm iterates the following two steps;
\begin{enumerate}
\item for each trajectory $\y^k$, for all $h^k = (z^k, t_s^k, t_e^k)$ the modified Kalman filter is paplied to estimate $\{ \hat\x^k \}$ and $\gamma^k$.
\item Update $\Theta$.
\end{enumerate}

\def\srho{\rho}
\let\oldrho\rho
\def\rho{\boldsymbol{\oldrho}}
\def\A{\boldsymbol{A}}
\def\B{\boldsymbol{B}}
\def\X{\boldsymbol{X}}
\def\Y{\boldsymbol{Y}}
\def\Z{\boldsymbol{Z}}
\def\v{\boldsymbol{v}}

\subsection{HMM and Switching Kalman Filter}

\begin{figure}[tb]
\begin{center}
\includegraphics[width=0.8\linewidth]{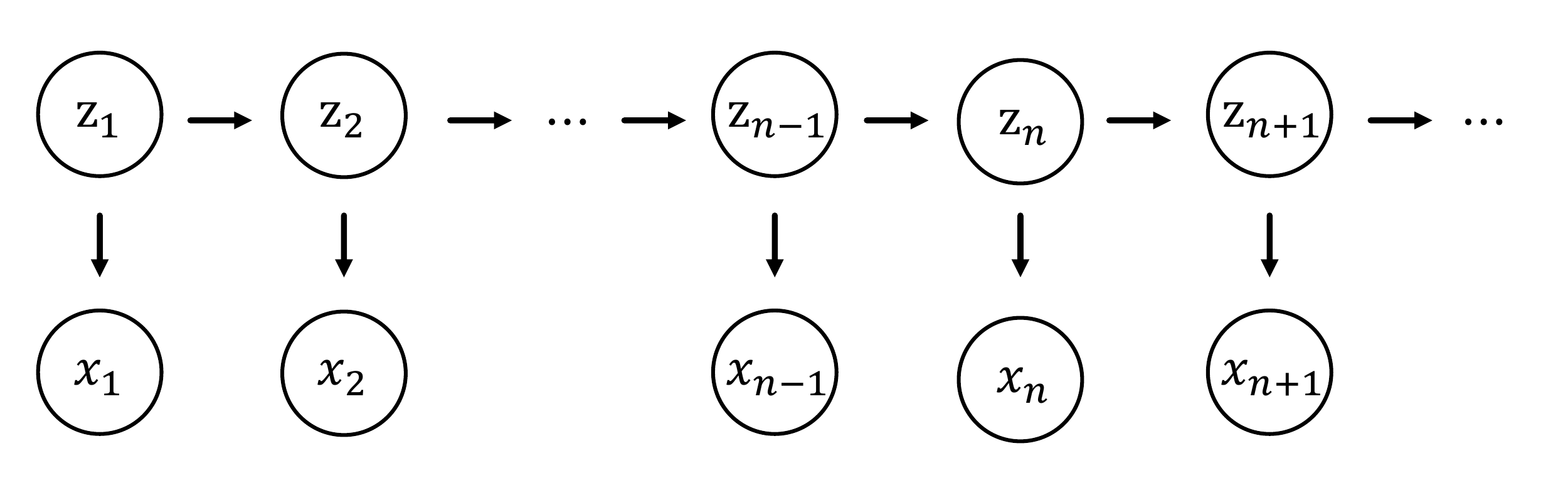}
\end{center}
\caption{State space model of HMM}
\label{fig:2}
\end{figure}

A Hidden Markov Model (HMM) shown in Figure \ref{fig:2} has  discrete latent variables $z$.
By using the Baum-Welch algorithm \cite{Baum}, HMM learns parameters from training data,
then infers unobserved states $\Z = \{z_n\}_{n=1}^N$ from observations $\X = \{\x_n\}_{n=1}^N$
by using the Viterbi algorithm \cite{Viterbi}.

\begin{figure}[tb]
\begin{center}
\includegraphics[width=0.8\linewidth]{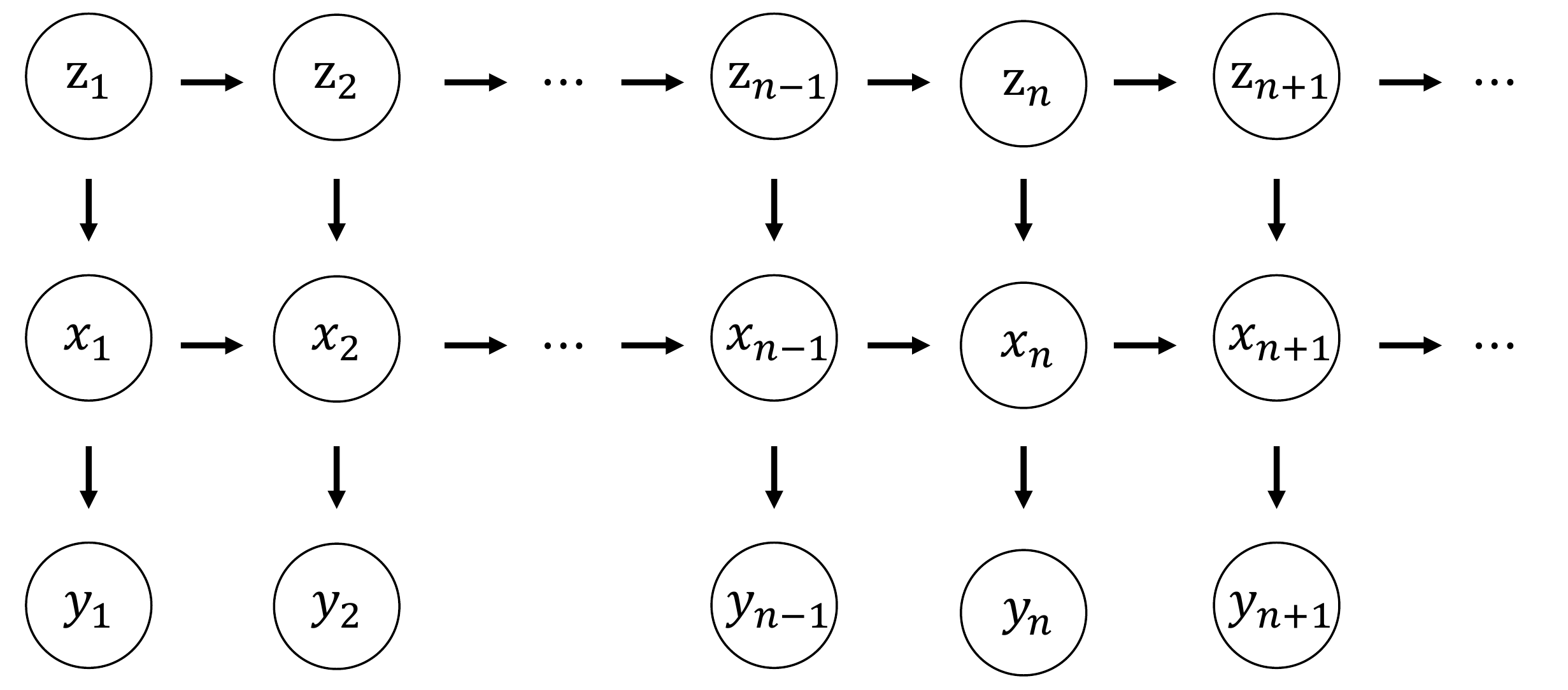}
\end{center}
\caption{State space model of Switching Kalman Filter}
\label{fig:3}
\end{figure}

A possible extension of MDA to segmentation is shown in Figure \ref{fig:3}
where $z_n$ is assigned to each state $\x_n$ by using HMM,
where $\x_n$ is a state of Kalman filter,
and $z_n$ is a hidden variable indicating which agent the observation is generated.
Both $\x_n$ and $z_n$ are dependent on previous variables $\x_{n-1}$ and $z_{n-1}$
as in Fig. \ref{fig:3}, which is known as Switching Kalman Filter \cite{murphy1998,bar1993}.

Switching Kalman Filter \cite{murphy1998,bar1993}
is a dynamic system model having parameters that depend on hidden variables.
State $\x_n$ and observation $\y_n$ at time $n$ is given by
\begin{align}
\x_n &\sim P(\x_n | \x_{n-1}) = N(\x_n | A_n \x_{n-1}, Q_n)
\\
\y_n &\sim P(\y_n | \x_n) = N(\y_n | C_n \x_n, R_n)
\end{align}
where $A_n = A[z_n]$，$C_n = C[z_n]$，$Q_n = Q[z_n]$，$R_n = R[z_n]$ are parameters
that are switched by the value of hidden variable $z_n$.

Switching Kalman Filter is a useful model, however
needs state transition probabilities to be given \cite{bar1993},
therefore is not applicable to the task presented here.
We instead propose to separate MDA agent estimation from HMM inference
to make the whole procedure to work.

\begin{figure*}[t]
\begin{center}
\includegraphics[width=0.8\linewidth]{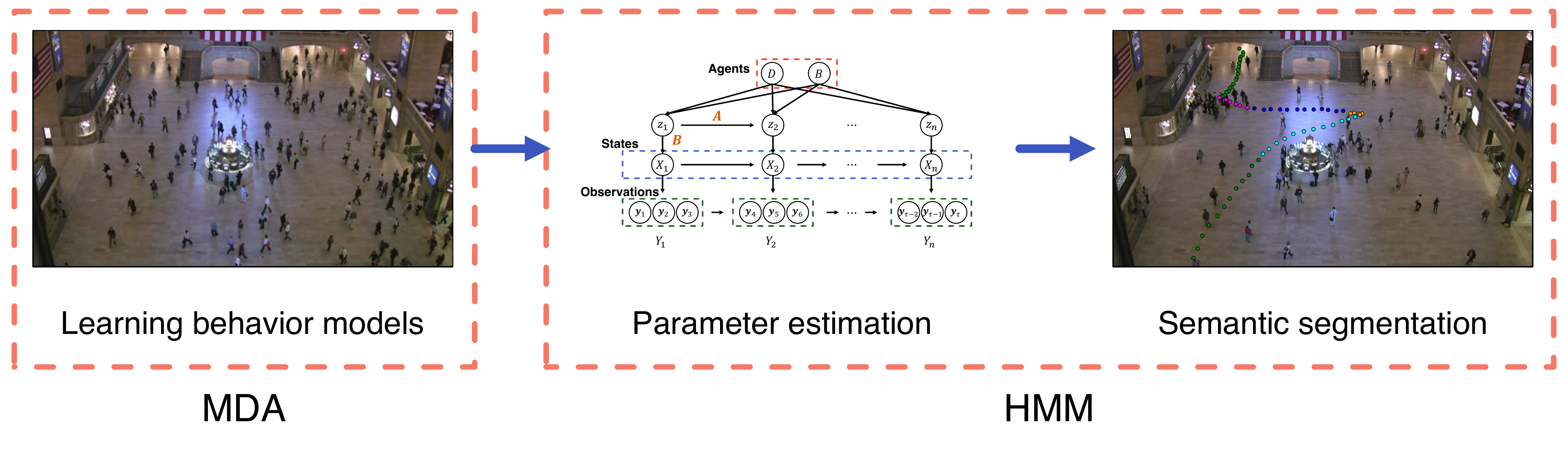}
\end{center}
\caption{Overview of the proposed method}
\label{fig:4}
\end{figure*}

\section{Proposed method}

Figure \ref{fig:4} shows the overview of the proposed method.
First we learn multiple agent models of trajectories from videos by using MDA.
Then we segment trajectories by using HMM based on the learned agents.

\subsection{Agent estimation by MDA}
\label{subsec:MDAinProposedmethod}

Let $M$ agents be $D_m = (A_m, \b_m, Q_m, R_m)$, 
$B_m = (\mu_{s,m}, \Phi_{s,m}, \mu_{e,m}, \Phi_{e,m})$,
and $\pi_m$.
Then all agents are denoted as 
\begin{align}
\Theta = \{ (D_m, B_m, \pi_m)\}_{m=1}^{M} = \{\omega_m\}_{m=1}^{M}
\end{align}
and these are estimated as shown section \ref{subsec:MDA}.

\subsection{HMM parameter estimation}
\label{subsec:HMMbaum}

\begin{figure}[tb]
\begin{center}
\includegraphics[width=\linewidth]{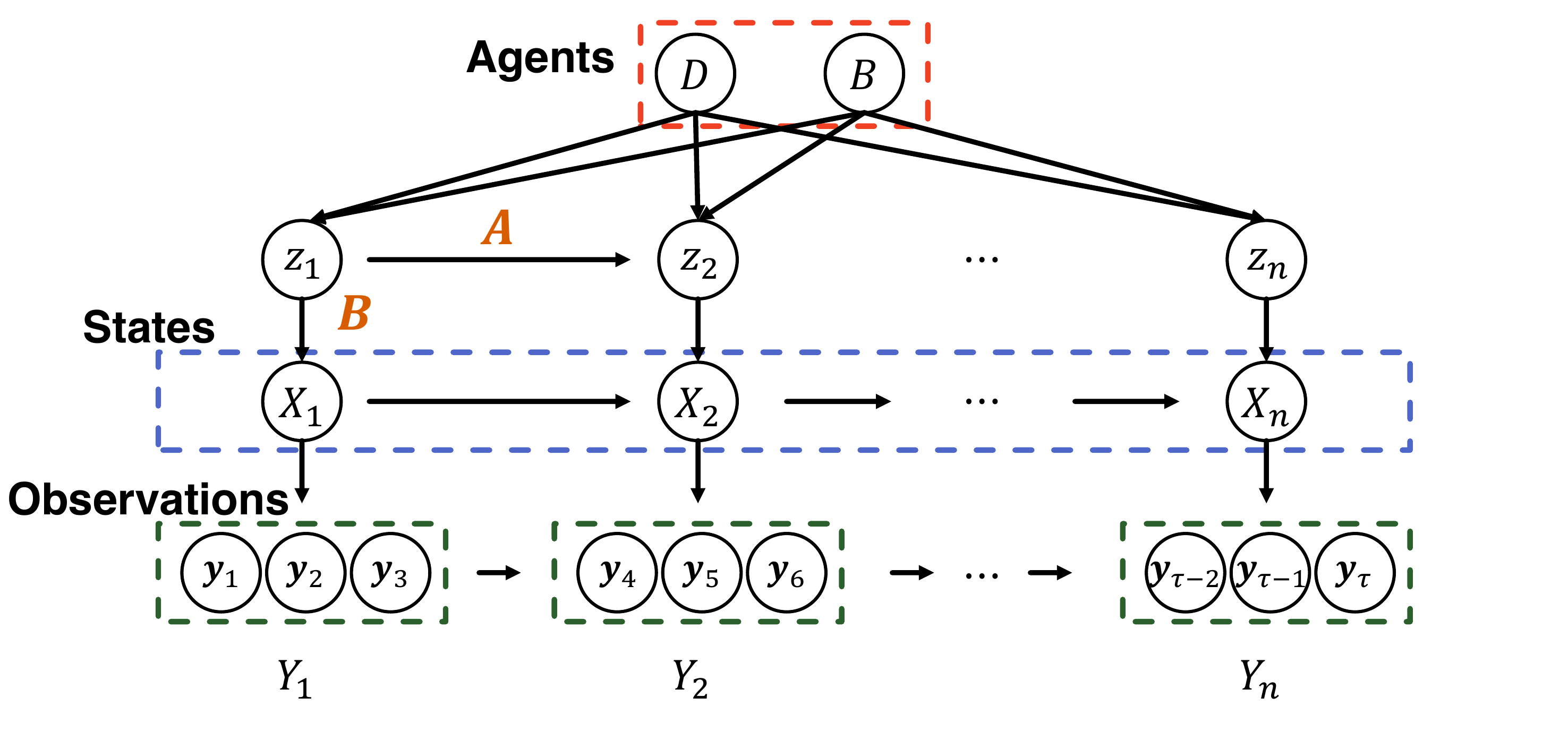}
\end{center}
\caption{State space model of the proposed method}
\label{fig:5}

\end{figure}

Figure \ref{fig:5} shows the model of the proposed method.
Agents transit from one another according to state transition matrix $\A$,
and state $\X$ is generated based on output probability matrix $\B$.
We use the Baum-Welch algorithm \cite{Baum}
to estimate initial probability distributions $\rho$
of learned $M$ agents, as well as matrices $\A$ and $\B$.


We assume that an agent may switch to other agent at each step,
and observation $\Y_t \in R^6$ of one step consists of successive three coordinates 
$\y_{t_1}, \y_{t_2}, \y_{t_3} \in R^2$
in a trajectory, as shown in Fig. \ref{fig:5}.
State $\X_t$ is considered to be generated
by an agent specified $z_t$ associated to step $t$.
Hence a trajectory is represented by 
hidden variables $\Z = \{z_t\}_{t=1}^{n}$,
states $\X = \{\X_t\}_{t=1}^{n}$,
and observations $\Y = \{\Y_t\}_{t=1}^{n}$.

Here let $\rho$ be an $M$-dimensional vector
whose $m$-th element represents 
the initial distribution $\srho_m$ of agent $\omega_m$,
and $\A$ by an $M\times M$ matrix
whose $(i, j)$ element is transition probability $a(i, j)$
from agent $\omega_i$ to agent $\omega_j$.
Output distribution of agent $\omega_m$ is assumed to be normal $N(\mu_m, \Sigma_m)$,
and let $\B$ a vector whose $m$-th element is
output probability of agent $\omega_m$.
Also we denote probability that agent $\omega_j$ outputs state $\X_t$ by
\begin{align}
b(j, t) \sim p(\X_t| \omega_j)= N(\X_t| \mu_j, \Sigma_j)
\end{align}


Denoting HMM parameters $\rho, \A, \B$ to be estimated by $\Theta = (\rho, \A, \B)$,
we maximize the following log likelihood to estimate 
$\Theta$ given $K$ trajectories;
\begin{align}
Q(\Theta, \Theta^{old}) = \sum_{K}^{} \sum_{\Z}^{} p(\Z| \X, \Theta^{old})\ln{p(\X, \Z; \Theta)}
\end{align}
by using the EM algorithm.

\subsection{Semantic segmentation}
\label{subsec:HMMviterbi}

A trajectory is segmented by applying Viterbi algorithm \cite{Viterbi} with the learned HMM parameters 
$\Theta$, that is, sequences of hidden variables $\Z^{\ast}$
and agents $\boldsymbol{\Omega}^{\ast}$;
\begin{align}
&\Z^{\ast} = \bigl\{ i_1, i_2, \cdots, i_n \bigr\}
\\
&\boldsymbol{\Omega}^{\ast} = \bigl\{ \omega_{i_1}, \omega_{i_2}, \cdots, \omega_{i_n} \bigr\}
\end{align}

\section{Experiments}

We compare the proposed method, denoted by MDA+HMM in the following, with
the Ramer-Douglas-Peucker (RDP) algorithm \cite{Ramer,Douglas-Peucker}
in terms of segmentation accuracy.
Trajectories in the Pedestrian Walking Path Dataset \cite{yi2015}
are used for this experiments. This dataset has a large number of pedestrian trajectories
in videos of size $1920 \times 1080$ pixels.
First we evaluate methods with synthetic trajectories generated from the dataset
for performance comparison,
then with real trajectories of the dataset.

\subsection{Metrics}
\label{subsec:metrics}

Evaluation metrics used in this experiments 
are Positional error and Step error defined in Algorithm \ref{alg1}.
Note that $N_{est}$ and $N_{gt}$ are numbers of 
estimated and actual segmentation points in a trajectory.

\newcommand{\argmax}{\mathop\mathrm{argmax}\limits}
\newcommand{\argmin}{\mathop\mathrm{argmin}\limits}

\begin{algorithm}

\caption{Calculate positional and step errors}
\label{alg1}

\begin{algorithmic}[1]
\Function{CalcError}{$S_{1}, S_{2}$}
\State $pos = stp = 0$
\For{i}
\If{$S_{1}(i)$ is a Seg point} 
\State $\hat{j} = \argmin_{S_{2}(j)\ \text{is a segmentation point} } |j - i|$
\State $pos += \|obs(\hat{j}) - obs(i)\|$
\State $stp += |\hat{j} - i|$
\EndIf
\EndFor
\State \Return $pos, stp$
\EndFunction
\State $pos, stp = \Call{CalcError}{est, gt} + \Call{CalcError}{gt, est}$
\State $pos, stp /= (N_{est} + N_{gt})$

\end{algorithmic}
\end{algorithm}

\subsection{Synthetic data}
\label{subsec:syntheticdata}

In order to compare methods with a large number of trajectories,
we generate 20,000 trajectories from MDA agent models learned from the dataset.
By assuming that transition probabilities are uniform,
these trajectories are sampled from the linear system of Eqs. (\ref{eq:xt}) and (\ref{eq:yt}).
In the following, we use 10,000 trajectories for HMM training (parameter estimation),
and the other 10,000 trajectories for HMM inference (segmentation).
Segmentation points are ones where agent models are switched from one another.

For the RDP method, we segment trajectories by changing the parameter values $\epsilon$,
then choose the best one.
In this case, $\epsilon = 69$ and $\epsilon = 80$ minimize each error.


For the proposed method, we choose different number of agents for segmentation (between 5 and 10)
for HMM parameter estimation and inference. 
Because 10 agents were learned by MDA, we perform the same procedure for 10 times (except the case of using all 10 agents) then report averaged results.

Table \ref{table:1} show comparison results.
The proposed method works better when the number of used agents is larger than eight.

\begin{table}[tb]
\caption{Results of the experiments using synthetic trajectories}
\label{table:1}

\begin{center}
\begin{tabular}{c|ccc}
Method & \#Agent & Positional error & Step error \\
\hline
 & 5 & $44.99\pm 6.25$ & $2.35\pm0.17$\\
 & 6 & $38.81\pm 2.70$ & $2.10\pm0.18$\\
MDA+HMM & 7 & $33.76\pm 3.01$ & $1.81\pm0.13$\\
(Ours) & 8 & $30.91\pm 4.09$ & $1.57\pm0.13$\\
 & 9 & $25.59\pm 3.34$ & $1.32\pm0.10$\\
 & 10 & $\bf{20.88}$ & $\bf{1.09}$\\
\hline 
 & $\epsilon$ & & \\
RDP & 69 & $33.69$ & $1.84$\\
 & 80 & $34.17$ & $1.82$\\
\end{tabular}
\end{center}
\end{table}

\subsection{Real data}\label{subsec:actualdata}

For evaluating methods with the real dataset,
we selected and manually annotated 104 trajectories
so that trajectories are segmented at the point where pedestrians turn their walking directions.

For the RDP method, we segment all trajectories by changing the parameter values $\epsilon$,
then choose the best one.
In this case, $\epsilon = 38$ and $\epsilon = 29$ minimize each error.


For the proposed method, we choose different number of agents for segmentation (between 5 and 10),
as we did in the previous section.
For separating the dataset for HMM parameter estimation and inference,
we perform four-fold cross validation.

\begin{figure*}[tp]
\centering
\begin{minipage}[t]{\textwidth}
\centering
\subfigure[]{\includegraphics[width=0.3\linewidth]{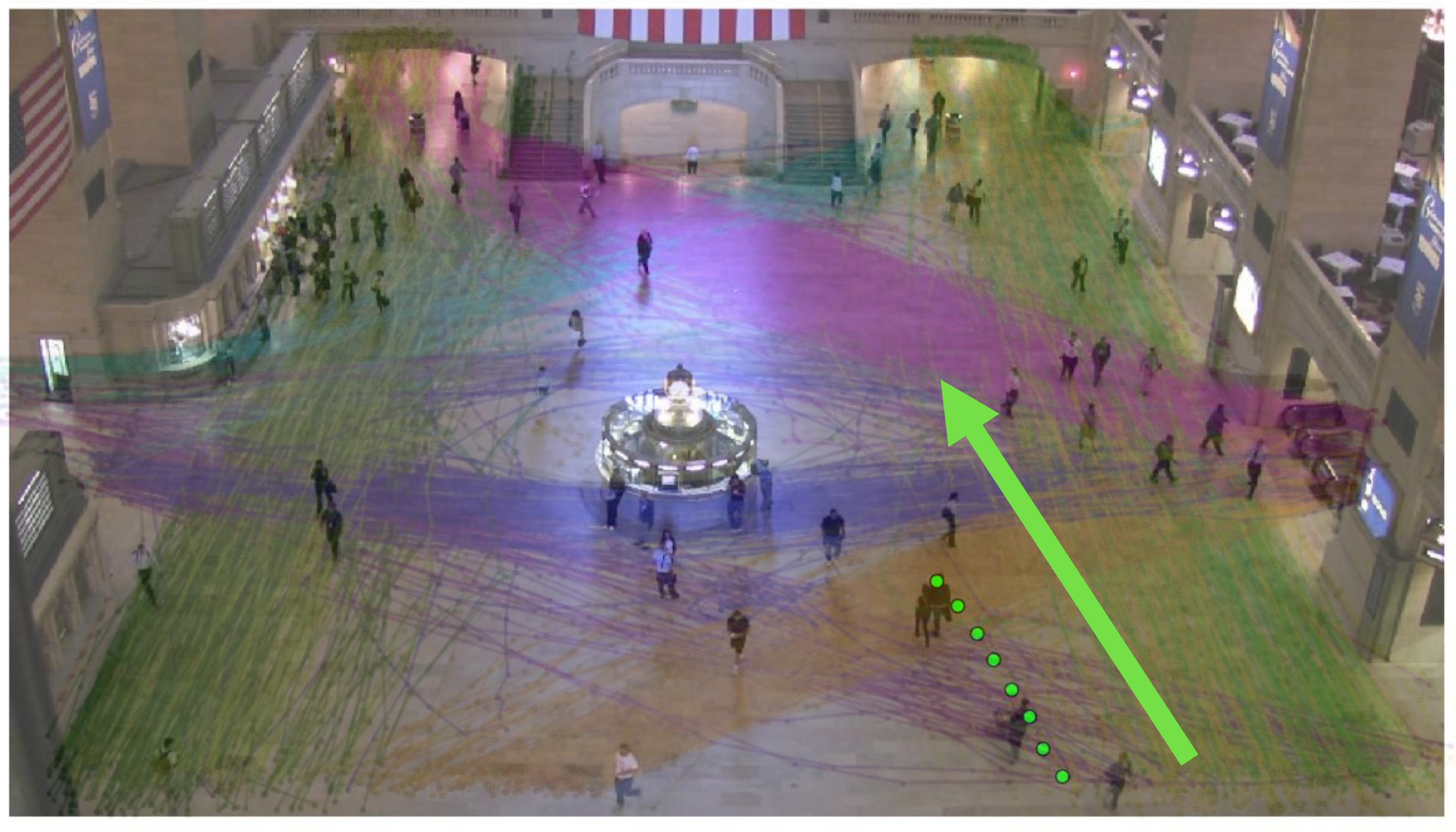}\label{fig:ped1-1}}%
\subfigure[]{\includegraphics[width=0.3\linewidth]{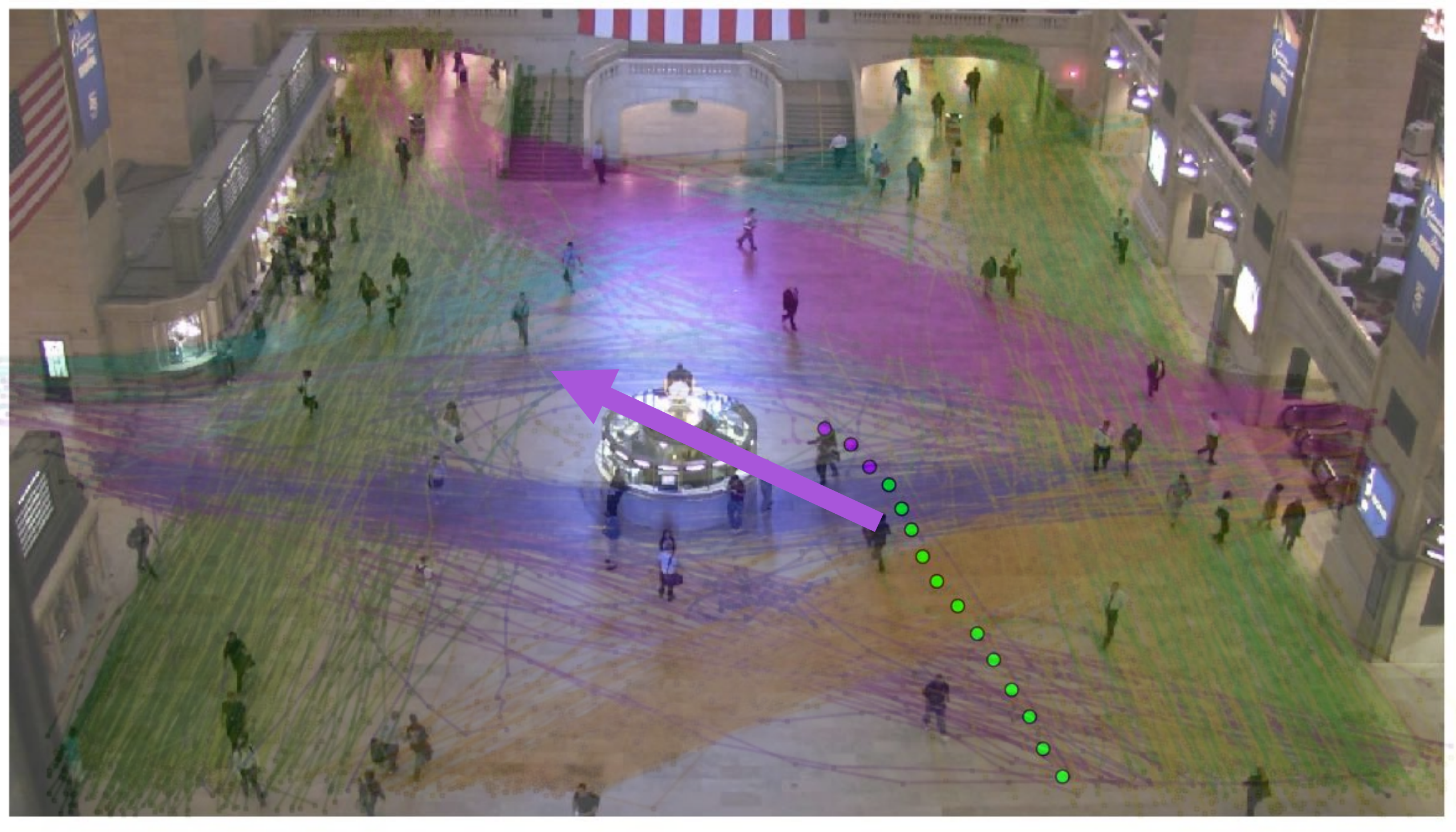}\label{fig:ped1-2}}%
\subfigure[]{\includegraphics[width=0.3\linewidth]{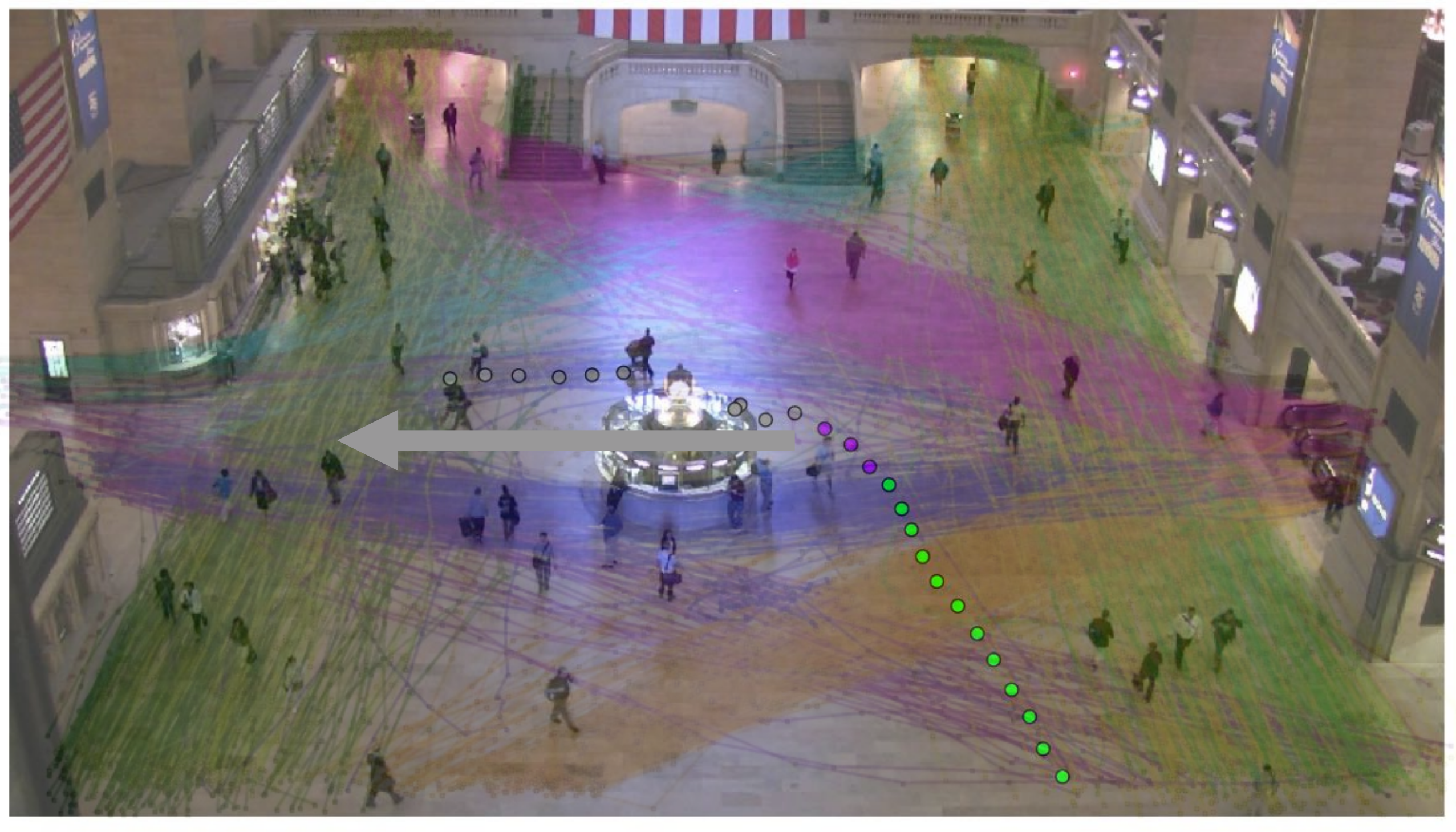}\label{fig:ped1-3}}%
\end{minipage}
\caption{Segmentation of pedestrian 1. The arrow indicates the waling direction. 
(a), (b), and (c) indicate temporal order; in (a) the pedestrian belongs to
an agent model going from right-bottom toward upper side,
in (b) to a model toward upper-left,
and in (c) to a model going left-side.}
\label{fig:ped1}
\end{figure*}

\begin{figure*}[tp]
\centering
\begin{minipage}[t]{\textwidth}
\centering
\subfigure[]{\includegraphics[width=0.3\linewidth]{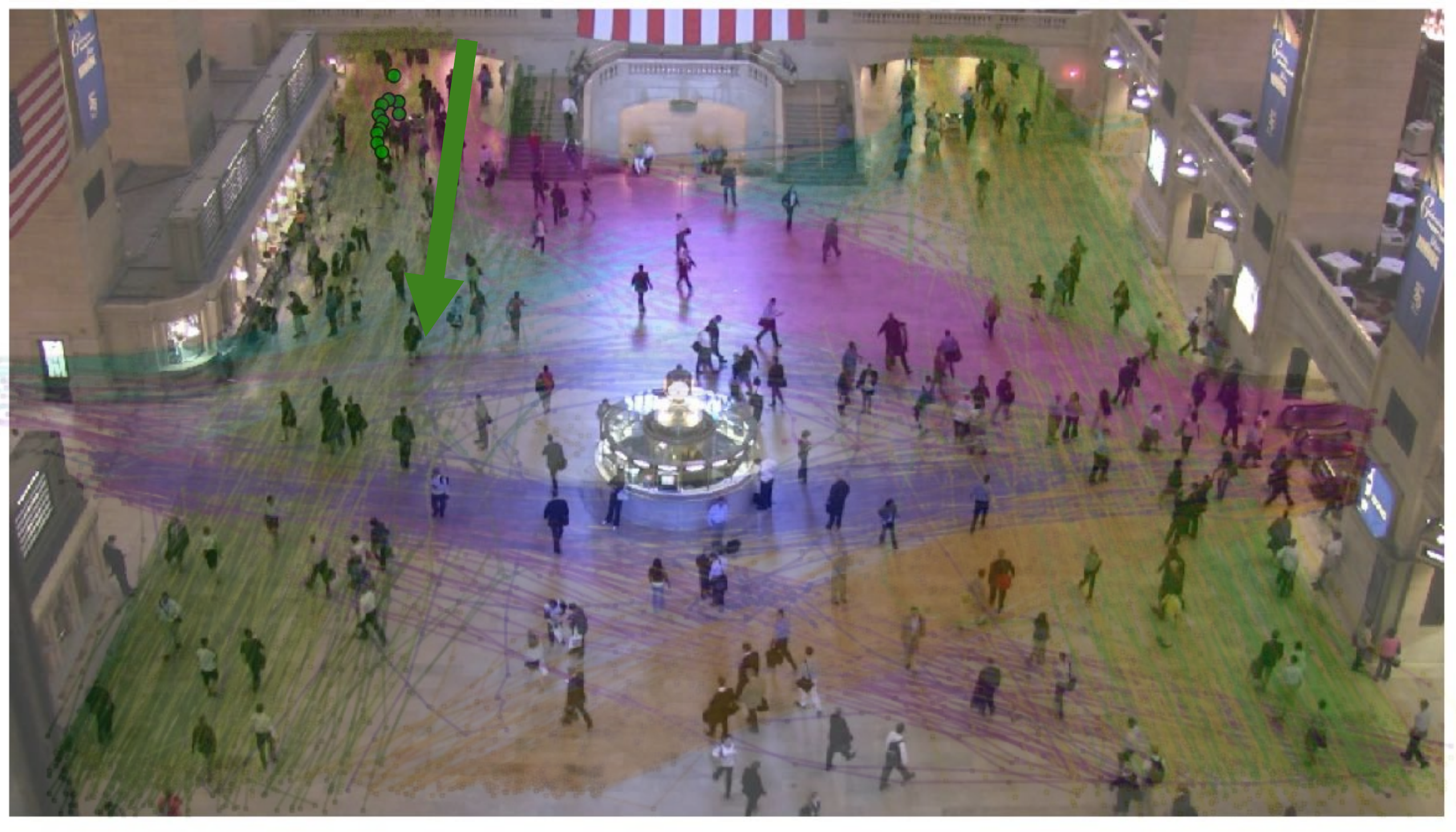}\label{fig:ped5-1}}%
\subfigure[]{\includegraphics[width=0.3\linewidth]{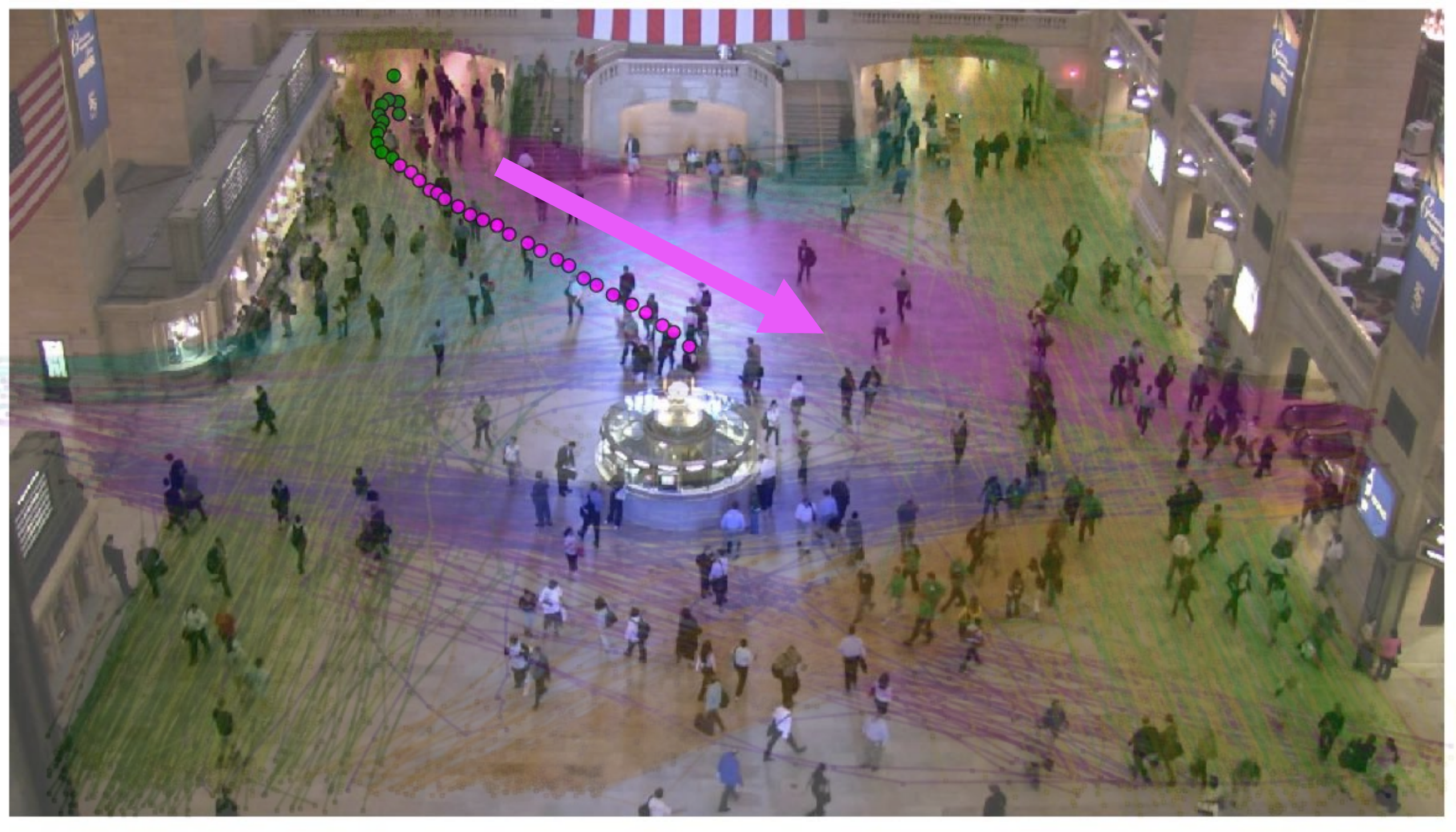}\label{fig:ped5-2}}%
\subfigure[]{\includegraphics[width=0.3\linewidth]{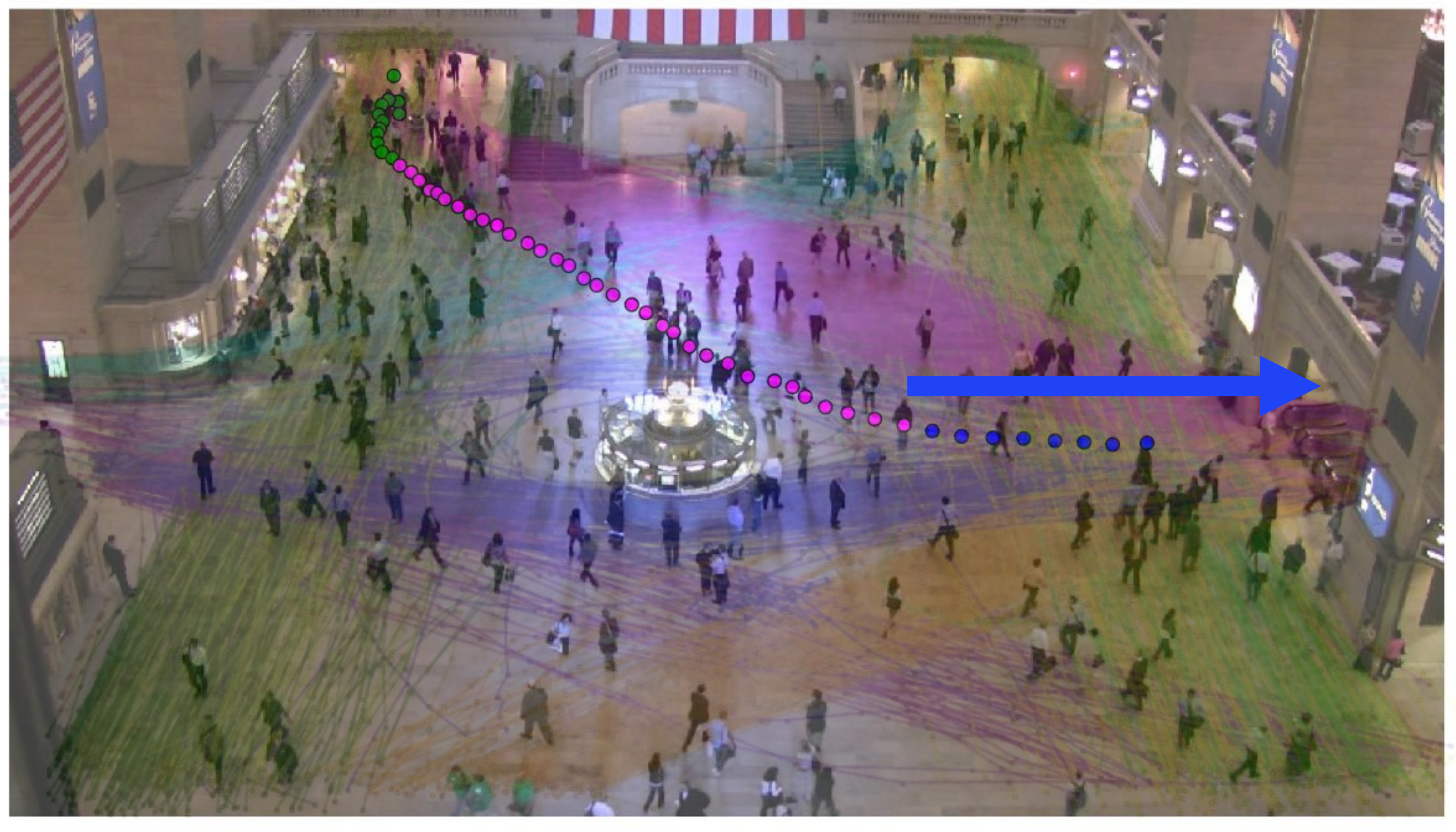}\label{fig:ped5-3}}%
\end{minipage}%
\caption{Segmentation of pedestrian 2. The arrow indicates the waling direction. 
(a), (b), and (c) indicate temporal order; in (a) the pedestrian belongs to
an agent model going downward,
in (b) to a model toward right-bottom,
and in (c) to a model going right-side.
}
\label{fig:ped2}
\end{figure*}

\begin{figure*}[tp]
\centering
\begin{minipage}[t]{\textwidth}
\centering
\subfigure[]{\includegraphics[width=0.3\linewidth]{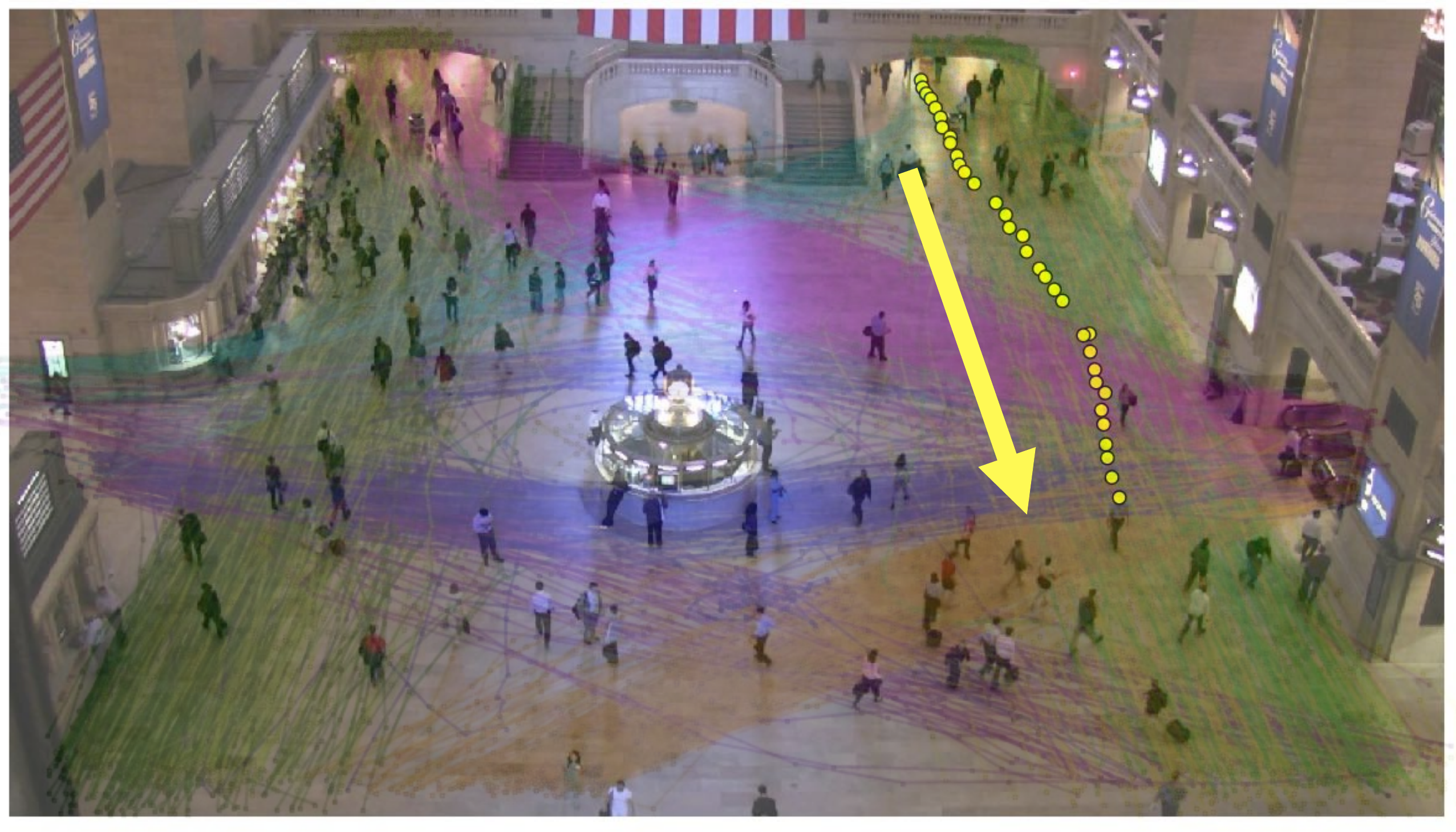}\label{fig:ped3-1}}%
\subfigure[]{\includegraphics[width=0.3\linewidth]{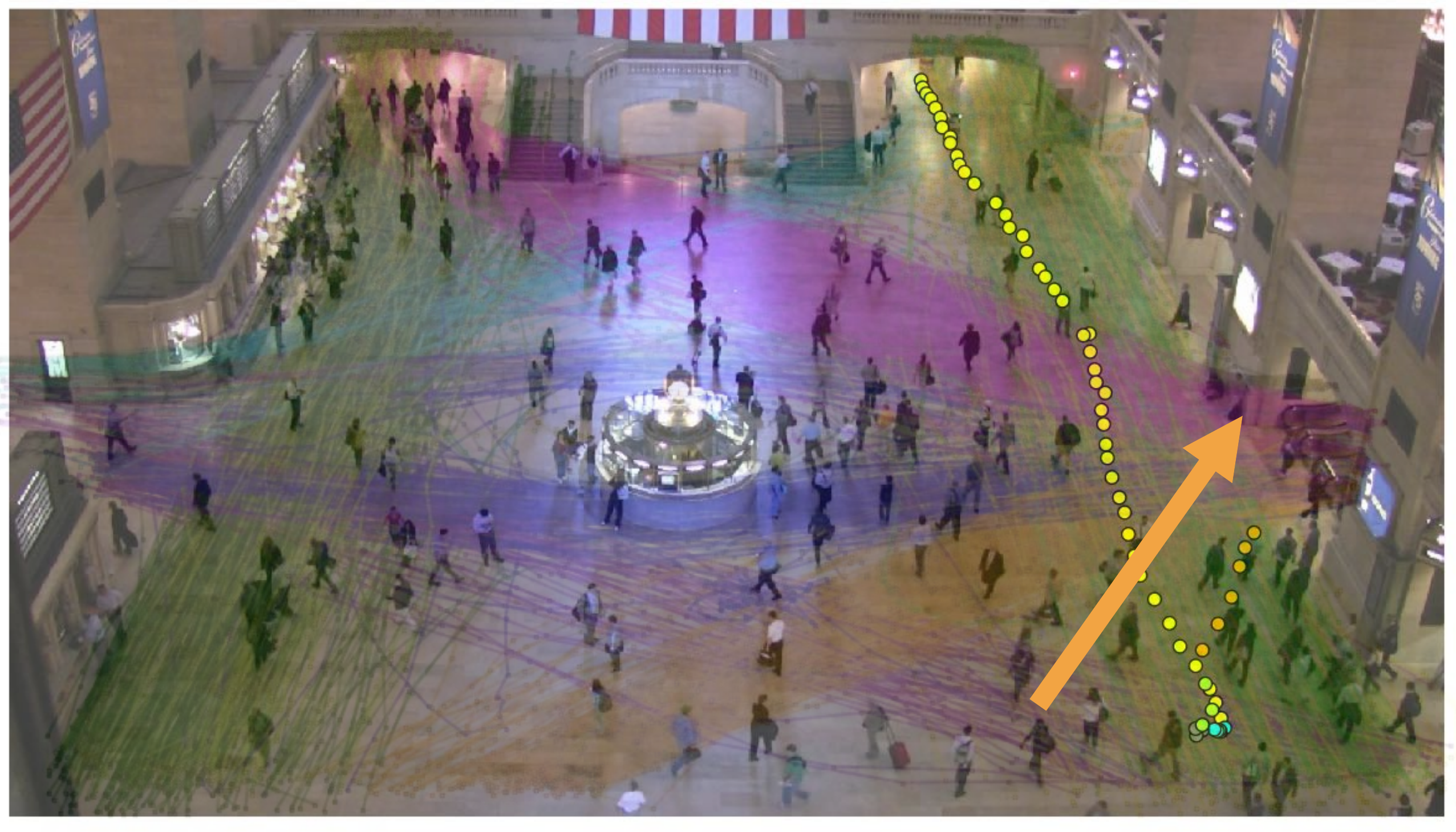}\label{fig:ped3-2}}%
\subfigure[]{\includegraphics[width=0.3\linewidth]{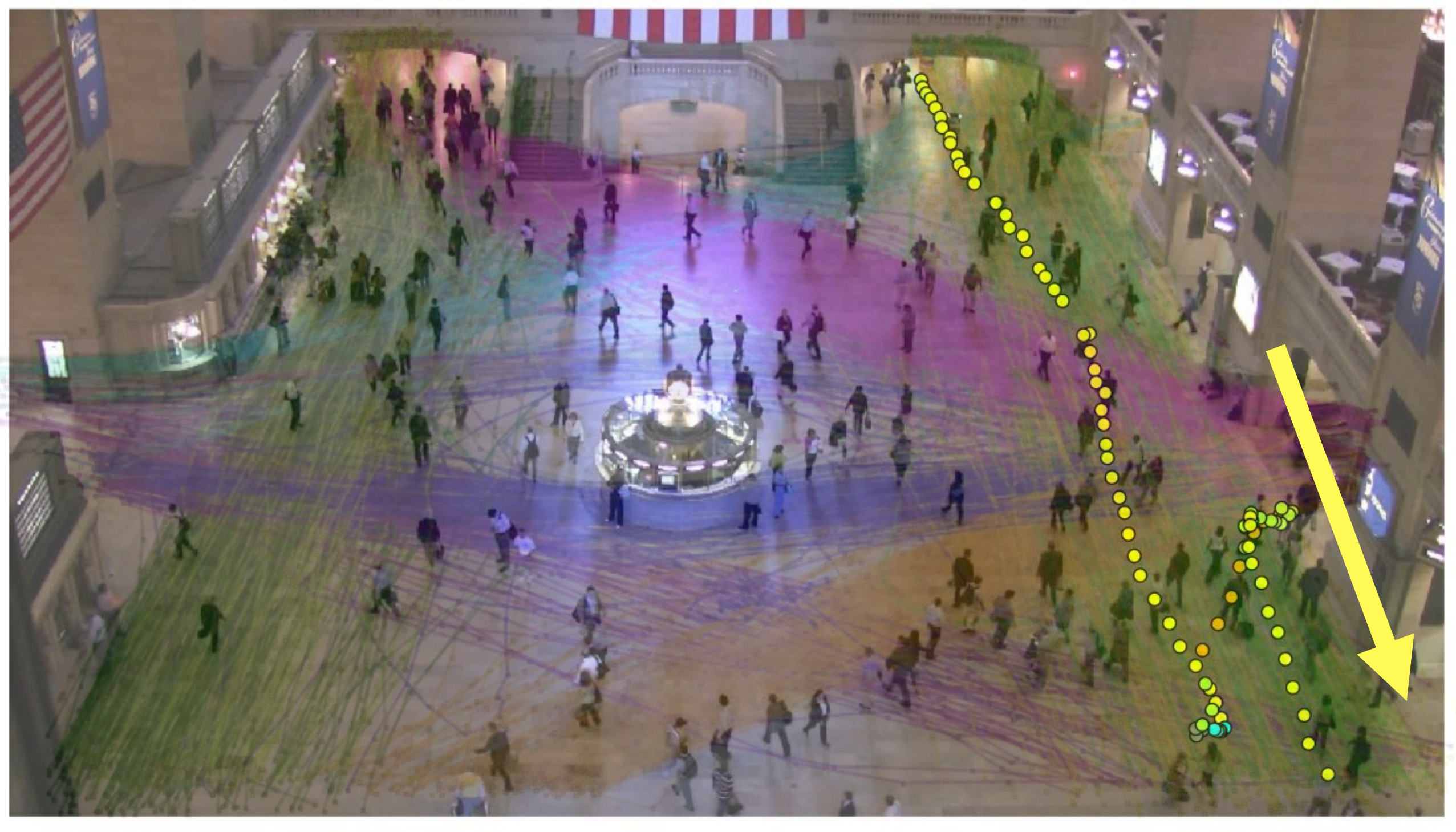}\label{fig:ped3-3}}%
\end{minipage}
\caption{
Segmentation of pedestrian 3. The arrow indicates the waling direction. 
(a), (b), and (c) indicate temporal order; in (a) the pedestrian belongs to
an agent model going downward,
in (b) to a model toward right,
and in (c) to a model going downward.
Notice that there exists many models around points where the pedestrian turns its direction abruptly.
}
\label{fig:ped3}
\end{figure*}

\begin{figure*}[tp]
\centering
\begin{minipage}[t]{\textwidth}
\centering
\subfigure[]{\includegraphics[width=0.3\linewidth]{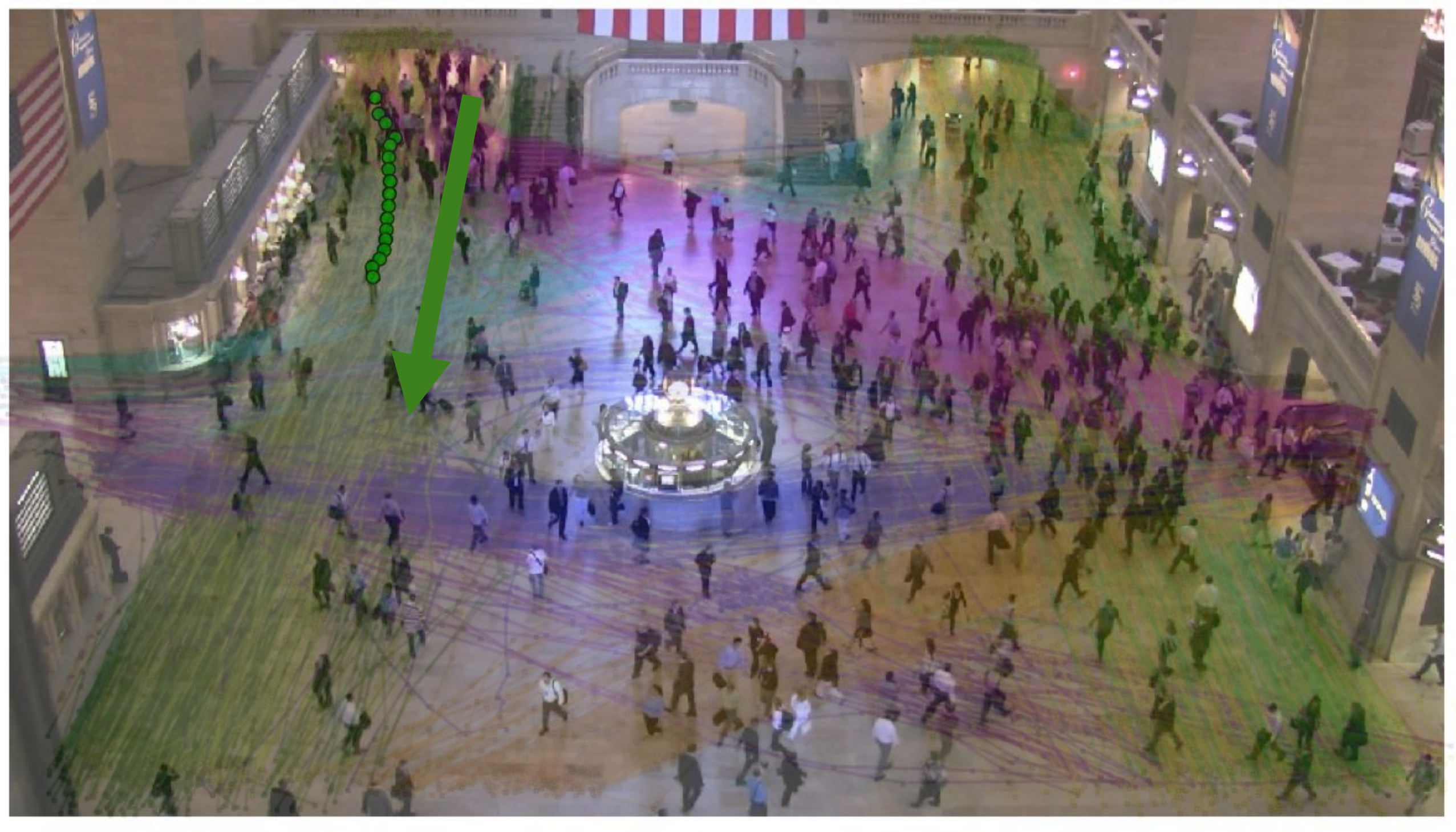}\label{fig:ped4-1}}%
\subfigure[]{\includegraphics[width=0.3\linewidth]{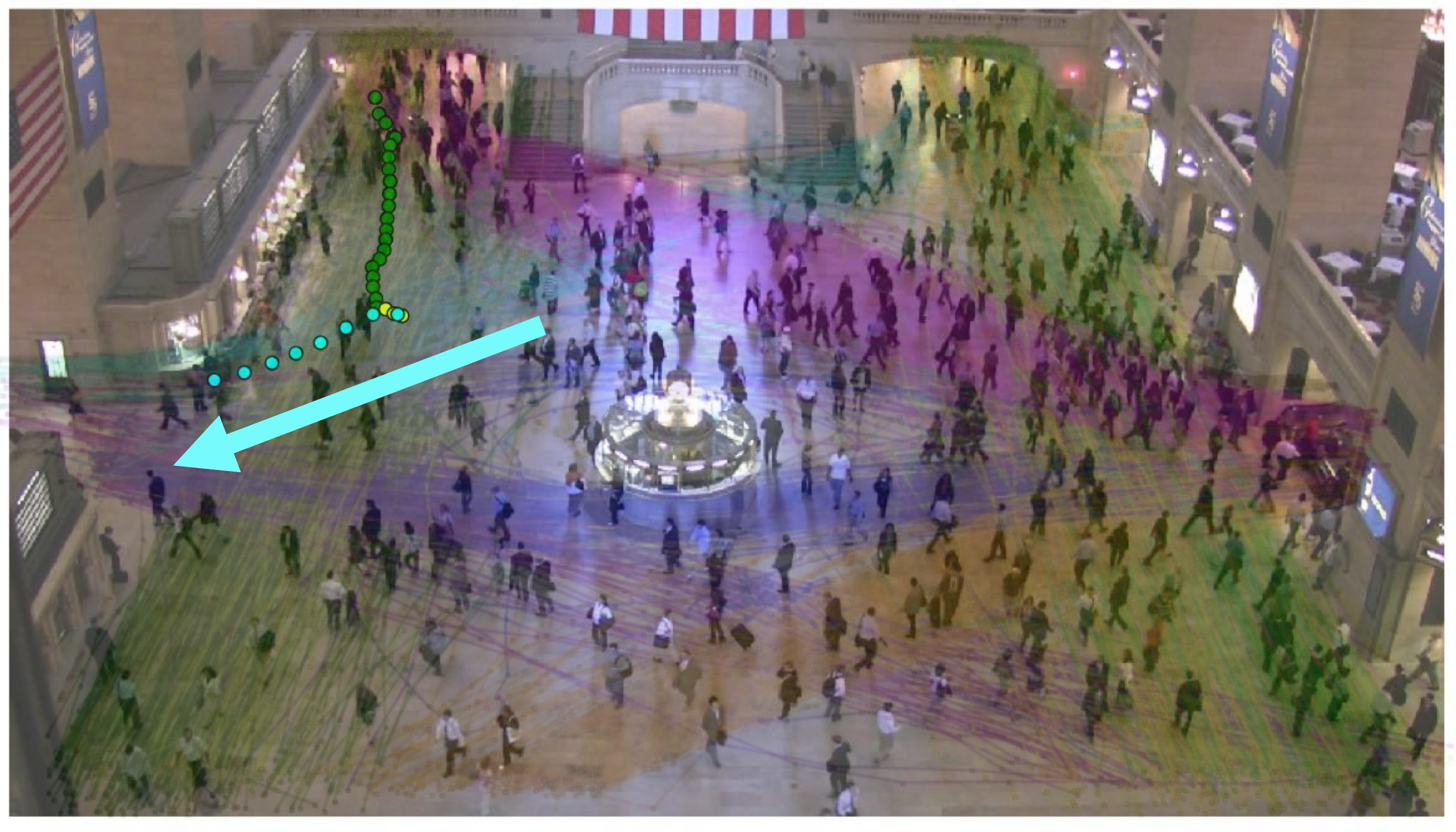}\label{fig:ped4-2}}%
\subfigure[]{\includegraphics[width=0.3\linewidth]{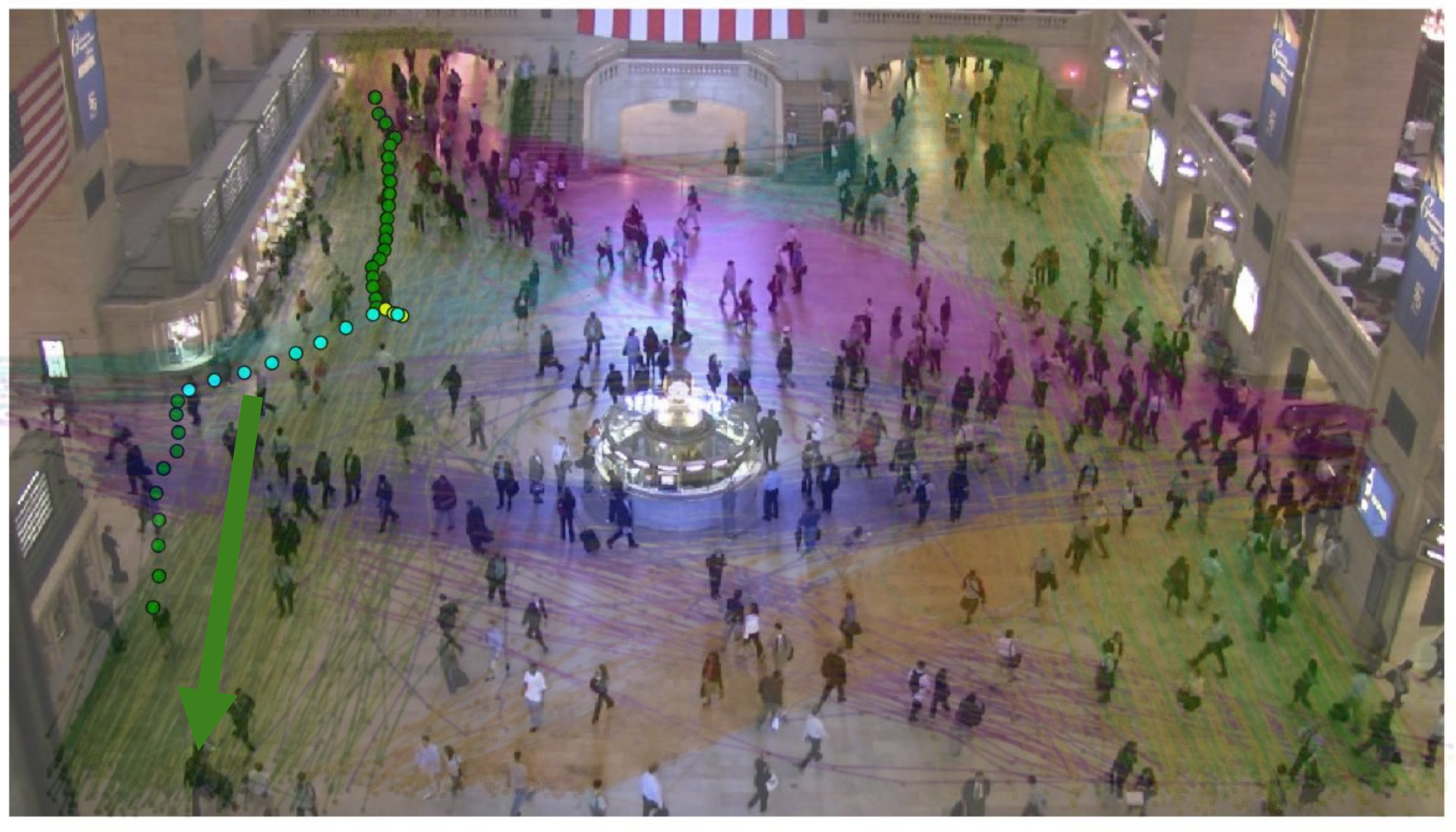}\label{fig:ped4-3}}%
\end{minipage}
\caption{
Segmentation of pedestrian 4. The arrow indicates the waling direction. 
(a), (b), and (c) indicate temporal order; in (a) the pedestrian belongs to
an agent model going downward,
in (b) to a model toward left-bottom,
and in (c) to a model going downward.
Notice that there exists many models around points where the pedestrian turns its direction abruptly.
}
\label{fig:ped4}
\end{figure*}

Results are shown in table \ref{table:2}.
Figures \ref{fig:ped1} to \ref{fig:ped4} visualize segmentation results
by the proposed method.
RDP errors are smaller then the proposed method, however
it doesn't provide any semantic information of segmentation.
In contrast, the proposed method divides trajectories
into semantically meaningful segments with associated agent models,
which helps to understand the behavior of the pedestrians in the real scene.

\begin{table}[tb]
\caption{Experimental result of using actual data}
\label{table:2}

\begin{center}
\begin{tabular}{c|ccc}
Method & \#Agent & Positional error & Step error \\
\hline
 & 5 & $62.12\pm 5.95$ & $2.92\pm0.27$\\
 & 6 & $57.69\pm 5.22$ & $2.70\pm0.20$\\
MDA+HMM & 7 & $56.41\pm 5.73$ & $2.61\pm0.25$\\
(Ours) & 8 & $\mathbf{52.32\pm 5.64}$ & $\mathbf{2.39\pm0.22}$\\
 & 9 & $53.69\pm 3.53$ & $2.46\pm0.15$\\
 & 10 & $53.41$ & $2.44$\\
\hline 
 & $\epsilon$ & &  \\
RDP & 29 & $27.85$ & $1.21$\\
 & 38 & $ 26.61$ & $1.24$\\
\end{tabular}
\end{center}
\end{table}

\section{Conclusions}

In this paper we have proposed a semantic trajectory segmentation method
by combining MDA and HMM to estimate agent models and segment trajectories
according to the learned agents.
Experimental results with synthetic trajectory dataset show that the 
proposed method works better than the baseline, the Ramer-Douglas-Peucker method.
Errors of the proposed method on the real dataset are relatively large
due to the fact that the HMM tends to infer multiple agents frequently at
turning points of pedestrians. Our future work includes to overcome this issue.


\ifCLASSOPTIONcompsoc
  \section*{Acknowledgments}
\else
  \section*{Acknowledgment}
\fi

This work was supported in part by JSPS KAKENHI grant number JP16H06540.



%



\bibliographystyle{IEEEtran}
\bibliography{bibtex_library}

\end{document}